\documentclass[10pt, a4paper]{article}

\usepackage[final]{lrec-coling2024} % this is the new style

\usepackage{adjustbox}
\usepackage{xcolor}
\newcommand{\red}[1]

\title{SciDMT: A Large-Scale Corpus for Detecting Scientific Mentions}

\name{Huitong Pan$^{\ast}$ Qi Zhang$^{\ast}$ Cornelia Caragea$^{\dagger}$ Eduard Dragut$^{\ast}$ Longin Jan Latecki$^{\ast}$} 

\address{$^{\ast}$Temple University \\ Philadelphia, Pennsylvania, USA\\
        \{huitong.pan, qi.zhang, edragut, latecki\}@temple.edu\\
        $^{\dagger}$University of Illinois Chicago \\ Chicago, Illinois, USA\\
            cornelia@uic.edu}

\abstract{
    We present SciDMT, an enhanced and expanded corpus for scientific mention detection, offering a significant advancement over existing related resources. SciDMT contains annotated scientific documents for datasets (D), methods (M), and tasks (T). The corpus consists of two components: 1) the SciDMT main corpus, which includes 48 thousand scientific articles with over 1.8 million weakly annotated mention annotations in the format of in-text span, and 2) an evaluation set, which comprises 100 scientific articles manually annotated for evaluation purposes. To the best of our knowledge, SciDMT is the largest corpus for scientific entity mention detection. The corpus's scale and diversity are instrumental in developing and refining models for tasks such as indexing scientific papers, enhancing information retrieval, and improving the accessibility of scientific knowledge. We demonstrate the corpus's utility through experiments with advanced deep learning architectures like SciBERT and GPT-3.5. Our findings establish performance baselines and highlight unresolved challenges in scientific mention detection. SciDMT serves as a robust benchmark for the research community, encouraging the development of innovative models to further the field of scientific information extraction.
 \\ \newline  }

\begin{document}

\maketitleabstract

\section{Introduction} \label{sec:intro}
% Challenges of SEMD algorithms and dataset.
Scientific entity mention detection (SEMD) is an instance of the Named Entity Recognition (NER) problem, which is usually a token-by-token tagging task. While NER has witnessed significant advancements owing to machine learning innovations~\citep{bose_2021}, SEMD remains in the early stages of exploration. The intricate and diverse terminologies used in scientific literature, coupled with the scarcity of extensively annotated corpora, exacerbate the complexity of SEMD.

% Why need a large and non-human-labeled dataset?
The existing corpora like RCC\footnote{https://github.com/Coleridge-Initiative/rclc}, SciERC, SciREX, and TDMSci \citeplanguageresource{bioNerDS,semeval2017,semeval2018,SCIERC,NLP-TDMS,SciRes,MDER,heddes2021,farber2021identifying} have been instrumental for SEMD algorithm evaluation but are constrained by their small volume and entity linking capabilities. These limitations stem from the manual curation process, which, while ensuring quality, is resource-intensive and scales poorly.

\begin{figure*}[t]
\includegraphics[width=1\linewidth]{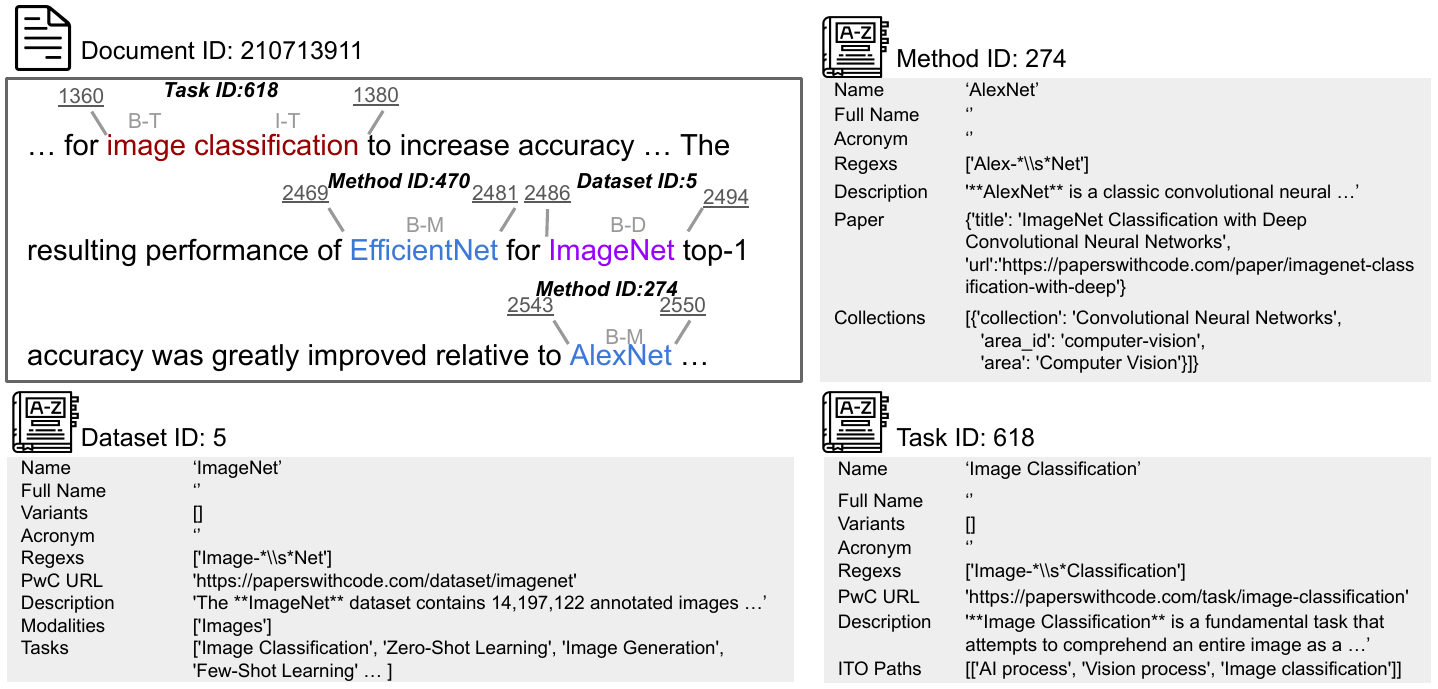}
\caption{Example document-level annotation (top-left) and dictionary entries in SciDMT. We mark each occurrence of \textbf{{\color[HTML]{9700FA}dataset (D)}}, \textbf{{\color[HTML]{3779D5}method (M)}} and \textbf{{\color[HTML]{9E0709}task (T)}} in papers and give the \underline{in-text spans}, \textit{entity indexes} and the BIO tags. For example, the method mention `EfficientNet' spans from 2469 to 2481 and has a BIO tag as `B-M'.}
\label{fig:example_ann} %Updated 06/15
\end{figure*}

% we introduced SciDMT and paper contribution
In this paper, we present SciDMT, a corpus featuring comprehensive entity annotations spanning datasets, methods, and tasks. SciDMT contains weakly labeled instances for model training and manually annotated instances for evaluation, offering a comprehensive resource for the advancement of SEMD.

The creation of SciDMT is facilitated by distant supervision~\cite{mintz}, leveraging document-level annotations from the Papers with Code\footnote{https://paperswithcode.com/} (PwC) website. This approach yields a main corpus comprising 48,049 machine-learning articles annotated with in-text spans, marking the mentions of datasets, methods, and tasks (DMT). Although distant supervision does not achieve the precision of manual annotations, the volume of data it generates is instrumental for training competitive models \cite{emonet,AlshehriSDO22,su2019,ZhangHVD18,zhang2019invest}.

Our contributions are multifaceted. SciDMT is more than a corpus; it's a resource for enhancing information extraction.
By annotating full articles and preserving the context of entity mentions, SciDMT aids in term disambiguation and enhances recognition accuracy. Every mention is linkable to PwC, and our introduction of ontology-linking for tasks and datasets further enriches the corpus’s utility. 

SciDMT is particularly valuable for indexing scientific papers, facilitating advanced information retrieval, and making scientific knowledge more accessible.
We validate SciDMT’s efficacy through experiments, showcasing its superiority in training SEMD models compared to existing corpora. Furthermore, our evaluation of NER methods, including SciBERT and GPT-3.5, on SciDMT demonstrates the intricate challenges and prospects of SEMD. The SciDMT corpus can be accessed at HuggingFace Hub\footnote{https://huggingface.co/datasets/jopan/SciDMT}. Our contributions can be summarized as follows:

\begin{itemize}
\item We introduce SciDMT, a SEMD corpus annotated at the document level, covering datasets, methods, and tasks. Each mention is linked to PwC and enriched with ontology-linking, offering a comprehensive resource for information extraction.
\item We compare SciDMT to existing corpora and demonstrate its effectiveness in training competitive SEMD models. 
\item We evaluate several NER methods, including SciBERT and GPT-3.5, on SciDMT, and discuss the unique challenges encountered in SEMD.
\end{itemize}

\section{Related Work}\label{sec:related}

The landscape of natural language processing (NLP) is rich with endeavors to transform unstructured and semi-structured text into structured insights through information extraction (IE). This endeavor has extended into the domain of scientific literature, leading to the development of specialized corpora tailored for extracting information from complex scientific texts~\citeplanguageresource{DMDD, SCIERC, scirex, tdmsci, heddes2021, NLP-TDMS, MDER, bioNerDS, farber2021identifying, semeval2017, semeval2018, tabassum-etal-2020-wnut, 10.1145/1401890.1402008}. These corpora, diverse in their scope, encompass a range of domains including AI, biomedical, and social sciences, and are enriched with a large variety of annotations, such as single scientific entity mention, citation, and coreference information.

In this study, we focus on publicly accessible and related corpora containing at least two scientific entity types for the detection of scientific entity mentions. Notable among these are SciERC~\citeplanguageresource{SCIERC}, SciREX~\citeplanguageresource{scirex}, and TDMSci~\citeplanguageresource{tdmsci}.

SciERC is a dataset of 500 annotated scientific abstracts containing mention spans, their types (Task, Method, Metric, Material, Other-ScientificTerm, and Generic), coreference information between mentions, and binary relations annotations. Building upon SciERC, SciREX extends the annotation scope to document-level annotations and encompasses multiple IE tasks, including mention detection for scientific entities (e.g., Dataset, Metric, Task, and Method), salient entity identification, and relation identification. 
TDMSci contains annotations for task, dataset, and metric entities on 2,000 sentences extracted from NLP papers. However, TDMSci does not include annotations for method mentions.

Despite these contributions, there remains a gap filled by our proposed SciDMT dataset. SciDMT aligns with SciREX in terms of named entity categories but stands out due to its large-scale corpus and comprehensive entity linking for each mention. This unique combination positions SciDMT as a significant advancement in the field of scientific information extraction.

\section{SciDMT Corpus}~\label{sec:SciDMT}
In this section, we describe the construction of SciDMT's main corpus and the human-annotated evaluation sets. We also present a comprehensive comparison between SciDMT and related corpora.
 
\subsection{SciDMT's Main Corpus}
We present the construction of our primary corpus in this subsection. Figure~\ref{fig:example_ann} is an illustrative example of a SciDMT data entry, which includes the parsed scientific article and the in-text annotation for scientific mentions.

\subsubsection{Data Collection}
Although our data collection methodology is similar to the one in DMDD~\citeplanguageresource{DMDD}
in that parsed articles from S2ORC~\citeplanguageresource{S2ORC} and document-level annotations from Papers With Code (PwC) are utilized, we significantly extend their distance supervision annotation. We extract publications' metadata of methods and tasks from PwC and dataset information directly from the paper’s PwC webpage. This process yields 48,049 matched papers between S2ORC and PwC, identified via their ArXiv IDs.

\subsubsection{Annotation with Distant Supervision}
Utilizing user-provided data from PwC, we aimed to align these entity names with their occurrences in the body of the articles. Strict matching, though generally effective, encounters challenges due to the occasional inconsistencies in entity naming conventions between PwC and authors (e.g., ‘k-Means’ vs. ‘k-Means Clustering’, ‘GoogLeNet’ vs. ‘GoogleNet’).

To address this, we developed a comprehensive DMT entity dictionary with regular expressions (regex), which enables us to accommodate variations in entity naming with approximate matching. These regex are not crafted for individual entities but are generated based on a universal set of rules, enhancing the scalability of our annotation process.

The regex creation rules can be summarized as follows. First, optional spaces and dashes are allowed between words. Second, acronyms are created for entity names with multiple words, but only when the original entity name is not already an acronym. Third, various common version names are considered. For example, if `v3.0' appears in the name, we allow matching mentions with `v3', and if `18' appears in the name, we allow matching mentions with `2018'. Fourth, verbs in different tenses and nouns in plural and singular forms are allowed. Lastly, the casing is ignored in regex, except for special cases such as the lowercase of the name being a common English word or the name being very short. As such, each entity name variation has one regular expression. Examples of regex can be seen in Figure~\ref{fig:example_ann}.

Our annotation process, though not aiming for optimal regex creation, is designed to obtain a substantial volume of weakly labeled data, instrumental for the effective training of NER models.

To enhance the entity linking capabilities of SciDMT, we integrated ontology paths from ITO~\citeplanguageresource{ITO}, which offers a structured hierarchy of AI tasks and datasets. In this integration, we mapped task and dataset entities from SciDMT to their corresponding elements within the ITO hierarchy. For task entities, we showed the complete hierarchy path in ITO, whereas for dataset entities, we showed the associated tasks. This method established relationships between entities reflecting their positions in the ontology's structure. For instance, datasets used in 'Image Classification' tasks are linked together, and tasks like 'Image Classification' and 'Image Segmentation' are connected as they both fall under the category of 'Vision Process'. This integration of ontology paths not only enhances the comprehensiveness of SciDMT but also enriches its usages for detailed entity analysis.

All 48,049 articles are annotated with the full text, section spans, and both document-level and in-text entity annotations. The annotations are indexed to the SciDMT entity dictionary, as illustrated in Figure~\ref{fig:example_ann}.

\subsubsection{Data Preprocessing} \label{sec:preprocessing}

% To address this, we have a second stage of preprocessing. 
In this phase, we employ a comprehensive approach, combining all regular expressions crafted from the SciDMT's dictionary. This exhaustive search is applied across all 48,049 articles, aiming to capture and annotate a broader spectrum of DMT entity mentions, thereby mitigating the issue of missing mentions.

\begin{table*}[ht]
\begin{adjustbox}{width=1\linewidth}
\begin{tabular}{rrrrrrrrrrr}
\hline
\textbf{} & \multicolumn{1}{c}{\textbf{Inst.}} & \multicolumn{1}{c}{\textbf{}} & \multicolumn{2}{c}{{\ul \textbf{Dataset}}} & \multicolumn{2}{c}{{\ul \textbf{Task}}} & \multicolumn{2}{c}{{\ul \textbf{Method}}} & \multicolumn{2}{c}{{\ul \textbf{All}}} \\
\textbf{Corpus} & \multicolumn{1}{c}{\textbf{Unit}} & \multicolumn{1}{c}{\textbf{\# Inst.}} & \textbf{\# M.} & \textbf{\# U.M.} & \textbf{\# M.} & \textbf{\# U.M.} & \textbf{\# M.} & \textbf{\# U.M.} & \textbf{\# M.} & \textbf{\# U.M.} \\ \hline
SciERC & abstract & 500 & 770 & 644 & 1,281 & 1,067 & 2,090 & 1,760 & 4,141 & 3,445 \\
SciREX & paper & 438 & 10,615 & 2,865 & 32,526 & 12,893 & 98,458 & 34,030 & 141,599 & 47,974 \\
TDMSci & sentence & 444 & 612 & 478 & 1,615 & 999 & 0 & 0 & 2,227 & 1,476 \\
SciDMT & paper & 48,049 & 449,798 & 10,807 & 647,360 & 7,850 & 733,728 & 16,579 & 1,830,886 & 34,648 \\ \hline
\end{tabular}
\end{adjustbox}
\caption{Summary of corpora for scientific entities mention detection.
}
\label{tab:summary_corpora}
\end{table*}

\subsection{Evaluation Sets with Human Annotations} \label{sec:gold-standard}
% How we created. Overview of the set
We manually annotated two sets of instances for evaluation purposes, one from SciDMT and the other from SciREX. The inclusion of SciREX serves a dual purpose: it not only facilitates a comparative analysis of dataset quality but also aids in assessing the complexity level of our dataset during experimental evaluations. These evaluation sets were manually annotated by two NLP researchers using brat rapid annotation tool~\cite{brat}. We aggregated the annotations by keeping only the mentions where both annotators agreed. 

For SciDMT evaluation set called \textbf{SciDMT-E}, annotators were tasked with 100 papers that were sampled with the most number of DMT mentions among the randomly sampled SciDMT's valid set. Additionally, we randomly selected 10 unseen entities from each DMT category and annotated the 256 sentences containing these unseen entities. Annotators were instructed to verify the detected mentions from SciDMT's main corpus and identify any missing mentions in each paper. To ensure accuracy, annotators were directed to search the PwC website and Google to confirm the DMT entities during the annotation process. Full annotation instructions are provided in HuggingFace Hub\footnote{https://huggingface.co/datasets/jopan/SciDMT/re
solve/main/SciDMT\%20Annotation\%20Guideline.pdf}.

We assessed the level of agreement between annotators using the relaxed span matches method, which considers a match when the entity mention spans from the annotators overlap. On SciDMT-E, the resulting Cohen Kappa 0.87 represents a substantial inter-annotator agreement~\cite{measure_agreement}. SciDMT-E contains 14,846 sentences with DMT entities, where 3,345 sentences contain dataset mentions, 11,124 sentences contain method mentions, and 5,899 sentences contain task mentions. The annotated mentions in SciDMT-E can be linked to 1,070 entities listed in SciDMT's dictionary. On average, each annotator required approximately 1 minute to annotate one sentence or 30 minutes to annotate one document.

When using SciDMT-E as ground truth and exact match for comparison, SciDMT's weak annotation obtains an F1 score of 61.9\%, precision of 50.8\%, and recall of 79.4\%. The recall rate indicates that the distantly-implied signals from PwC are able to capture 79.4\% of scientific entities in text. The low precision suggests that a significant portion of human-annotated mentions does not exactly match with the machine-annotated mentions. This observation is attributed to many weak labels failing to include the full entity name. For example, distant supervision may provide a partial annotation in sentences containing `KITTI 2012', tagging `KITTI' but ignoring the version part of the name, i.e., `2012'.

For SciREX evaluation set (SciREX-E), we used the same annotation guideline to annotate 10 papers for DMT entities. The Cohen Kappa 0.76 also indicates a substantial agreement between annotators. SciREX-E contains 2,207 sentences with DMT entities. Compared to SciREX-E, SciREX's original annotation obtains an F1 score of 65.4\%, precision of 77.4\%, and recall of 56.6\%. The low recall suggests that many mentions were missing in the original annotations. For example, the original annotation often misses the `ImageNet' mentions in phrases such as `ImageNet pre-trained model'.

\subsection{Comparison with Related Corpora}
\label{sec:linking_capability}
We compare SciDMT with three related corpora in terms of size (Table~\ref{tab:summary_corpora}) and quality. For each of the three scientific entity types, we give the total number of entity mentions (\# M.) and the total number of unique mentions (\# U.M.) for each corpus.

\subsubsection{Corpora Size}
SciDMT is larger than the related corpora, in terms of the number of instances (\# Inst. = 48,049), instance units (Inst. Unit = Paper), and the number of mentions (\# M. of All = 1,830,886). Having document-level annotations compared to sentence-level annotations, SciDMT allows a larger model input scope (e.g., sentence before and after the target sentence), allowing for richer contextual information. Furthermore, since entity mentions in SciDMT appear in multiple sentences across the 48K papers, it provides a diverse set of training data for NER and Entity Linking.
This is captured by comparing \# M. and \# U.M. in Table \ref{tab:summary_corpora}. A large number of unique mentions indicates a wide range of scientific entities captured in SciDMT, while the high total number of mentions contributes to the training of robust models by providing a variety of background semantics related to scientific entities. 

\subsubsection{Entity Linking Annotation} 
Entity linking is the task of associating mentions in the text with their corresponding entities in knowledge bases, such as Wikipedia and Papers with Code. In the case of scientific entity mentions, entity linking is crucial as it allows users to access the correct dataset, source code, and source papers for empirical studies. Since SciDMT is created based on Papers with Code, all entities mentioned in SciDMT have explicit links to the Papers with Code website and a unique identifier. Because of the incorporation of ITO paths, dataset and task entities have intra-entity annotation as well.

In contrast, the related corpora do not have linking information about their entities because the annotators were not instructed to provide the linking annotation. Our attempts to link the entities in the related corpora to knowledge bases, such as Papers with Code and the ACL Anthology, were largely unsuccessful due to several reasons: 

1) Their mentions include extra characters or text strings. For example, the mention `fine-tuned U-Net' includes the descriptive text `fine-tuned' for the method. 

2) Their mentions include more than one entity, for example, `ImageNet pretrained VGG-19'. 

3) Their mentions do not include the actual entity name, for example, `models' and `methods'. This is because some related corpora are annotated with pronominal reference to entities, as defined in ACE 2005~\cite{ace2005}. Pronominal reference is not helpful in linking mentions to scientific entities, especially when the corpora are not annotated on the paper level and the proper name reference is missing from the annotated instance. Within this characteristic group, there are also confusing mentions not using the most commonly used names or missing parts of the names. For example, `VGG' can denote many possible models, such as VGG-16 and VGG-19. This further points toward the value of including linking attributes in the annotation whenever possible, as done in our work.

\section{Experimental Setup}
The experiments are designed for the task of scientific entity mention detection (SEMD) with three primary objectives in mind: establishing baseline performance on SciDMT, gaining insights into the difficulty of SEMD, and evaluating the effectiveness of using SciDMT for training. The experiments focus on three categories of scientific entities: datasets, methods, and tasks (DMT).

\subsection{Baseline Models}~\label{sec:ner_models}
We formulate the task of SEMD as a single-sentence tagging task, and we include a diverse set of models as baselines in our evaluation, namely Conditional Random Fields (CRF), Bidirectional Long Short-Term Memory (BiLSTM), BERT~\citep{bert_base_cased}, SciBERT~\citep{beltagy-etal-2019-scibert} and GPT-3.5~\citep{instruct_gpt}. 

For CRF, BiLSTM, BERT, and SciBERT, we conduct training in 3 rounds using randomly shuffled training sets. For CRF, we incorporate features such as Part-of-Speech (POS) tags and keywords.

For BiLSTM, we evaluate two additional variations where the pre-trained embedding layer is initialized with either GLoVe (BiLSTM-G)~\citep{glove} or Word2Vec (BiLSTM-W)~\citep{word2vec}. Tokens that are not mapped with pre-trained embeddings are initialized with zeros. The embedding layer for all tokens is updated during training. To ensure a fair comparison, we set the embedding dimension to 300 for BiLSTM, BiLSTM-G, and BiLSTM-W.
For BiLSTM-G, we utilize the embedding trained on Wikipedia and Gigaword, covering 30,612 tokens, while 134,802 tokens are missing in the pre-trained embeddings. For BiLSTM-W, we use the embedding trained on Google News and convert 68,553 tokens, with 96,861 tokens missing. We notice that many scientific entity names are missing in the pre-trained embeddings. 

For BERT and SciBERT, we use the pre-trained weights of base-cased BERT~\citep{bert_base_cased} and scivocab-cased SciBERT~\citep{beltagy-etal-2019-scibert}. We keep the same hyperparameters for training the models as in the original SciBERT~\citep{beltagy-etal-2019-scibert}, except for the batch size, which is set to 16.

GPT-3.5 is included in our model selection because Large Language Models (LLMs) have demonstrated impressive natural language understanding capabilities, including the capability for entity recognition~\cite{Gutierrez2022ThinkingAG}. We use Spacy-LLM~\footnote{https://github.com/explosion/spacy-llm}, which is a Python package that combines the language processing library spaCy with LLM backends. In terms of model specifications, we use spacy.NER.v2 as task, `DATASET, METHOD, TASK' as labels, and OpenAI's gpt-3.5-turbo-0613 as the LLM backend. The input to the model consists solely of the sentence text. The model's output is in the span format, which we convert to token-level BIO labels for evaluation purposes. We only use the Spacy-LLM's zero-shot setting without any examples or label definitions. We acknowledge that more sophisticated model tuning and prompt engineering may yield improved performance; however, our focus here is on presenting baseline results.

\begin{table*}[]
\fontsize{10}{12}\selectfont
\centering
\begin{adjustbox}{width=1\textwidth}
\begin{tabular}{r|lll|lll|lll}
\hline
Subset & \multicolumn{3}{c|}{SciREX-E: All} & \multicolumn{3}{c|}{SciDMT-E: All} & \multicolumn{3}{c}{SciDMT-E: Unseen} \\ \hline
 & \multicolumn{1}{c}{F1} & \multicolumn{1}{c}{Precision} & \multicolumn{1}{c|}{Recall} & \multicolumn{1}{c}{F1} & \multicolumn{1}{c}{Precision} & \multicolumn{1}{c|}{Recall} & \multicolumn{1}{c}{F1} & \multicolumn{1}{c}{Precision} & \multicolumn{1}{c}{Recall} \\ \hline
CRF & .270 ± .000 & .172 ± .000 & .626 ± .000 & .455 ± .000 & .313 ± .000 & .832 ± .000 & .185 ± .000 & .107 ± .000 & .683 ± .000 \\
BiLSTM & .328 ± .004 & .238 ± .003 & .525 ± .007 & .520 ± .002 & .411 ± .003 & .708 ± .002 & .176 ± .017 & .137 ± .009 & .251 ± .048 \\
BiLSTM-G & .325 ± .007 & .235 ± .007 & .527 ± .005 & .526 ± .005 & .414 ± .007 & .721 ± .005 & .192 ± .039 & .147 ± .029 & .279 ± .060 \\
BiLSTM-W & .329 ± .003 & .238 ± .004 & .529 ± .002 & .523 ± .004 & .411 ± .006 & .719 ± .002 & .188 ± .023 & .137 ± .016 & .304 ± .045 \\
BERT & .480 ± .007 & .372 ± .010 & \textbf{.674 ± .005} & .643 ± .004 & .523 ± .007 & \textbf{.835 ± .006} & .747 ± .007 & .721 ± .011 & .776 ± .008 \\
SciBERT & .490 ± .003 & .388 ± .004 & .666 ± .006 & \textbf{.649 ± .001} & \textbf{.531 ± .002} & .833 ± .003 & \textbf{.763 ± .009} & \textbf{.737 ± .012} & \textbf{.792 ± .023} \\
GPT-3.5 & \textbf{.503 ± .000} & \textbf{.499 ± .000} & .506 ± .000 & .586 ± .000 & .672 ± .000 & .520 ± .000 & .484 ± .000 & .701 ± .000 & .370 ± .000 \\
\hline
Subset & \multicolumn{3}{c|}{SciDMT-E: Datasets} & \multicolumn{3}{c|}{SciDMT-E: Methods} & \multicolumn{3}{c}{SciDMT-E: Tasks} \\ \hline
\multicolumn{1}{l|}{} & \multicolumn{1}{c}{F1} & \multicolumn{1}{c}{Precision} & \multicolumn{1}{c|}{Recall} & \multicolumn{1}{c}{F1} & \multicolumn{1}{c}{Precision} & \multicolumn{1}{c|}{Recall} & \multicolumn{1}{c}{F1} & \multicolumn{1}{c}{Precision} & \multicolumn{1}{c}{Recall} \\ \hline
CRF & .590 ± .000 & .449 ± .000 & .858 ± .000 & .410 ± .000 & .276 ± .000 & .799 ± .000 & .393 ± .000 & .259 ± .000 & .813 ± .000 \\
BiLSTM & .551 ± .002 & .438 ± .004 & .743 ± .004 & .474 ± .003 & .363 ± .003 & .684 ± .001 & .489 ± .002 & .377 ± .003 & .696 ± .003 \\
BiLSTM-G & .558 ± .005 & .443 ± .007 & .756 ± .004 & .480 ± .007 & .365 ± .008 & .698 ± .003 & .496 ± .004 & .382 ± .006 & .706 ± .006 \\
BiLSTM-W & .552 ± .005 & .435 ± .005 & .755 ± .007 & .476 ± .005 & .363 ± .006 & .693 ± .000 & .492 ± .007 & .377 ± .008 & .708 ± .001 \\
BERT & .679 ± .004 & .550 ± .005 & \textbf{.886 ± .003} & .602 ± .005 & .478 ± .008 & .812 ± .004 & .560 ± .001 & .430 ± .005 & \textbf{.804 ± .013} \\
SciBERT & \textbf{.678 ± .003} & \textbf{.551 ± .002} & .881 ± .003 & \textbf{.611 ± .001} & .490 ± .001 & \textbf{.813 ± .003} & .565 ± .003 & .438 ± .004 & .795 ± .007 \\
GPT-3.5 & .663 ± .000 & .729 ± .000 & .608 ± .000 & .582 ± .000 & \textbf{.628 ± .000} & .543 ± .000 & \textbf{.579 ± .000} & \textbf{.620 ± .000} & .543 ± .000
\\ \hline
\end{tabular}
\end{adjustbox}
\caption{NER model performance on human-annotated evaluation sets. In each column, the highest score is shown in \textbf{boldface}.}
\label{tab:sotaner}
\end{table*}

\subsection{Train-Valid Split}
To establish a train-valid split for SciDMT's main corpus and SciREX, we first perform a random document-level split. Next, we randomly select 30 scientific entities, with 10 entities chosen from each DMT category in SciDMT. These entities form the unseen set, which is exclusively included in the valid set and the evaluation set, and is excluded from the training set of any corpus. Finally,  we conduct a sentence-level train-valid split based on the aforementioned document-level split.

At the document level, the train set of SciDMT consists of 36,635 documents (76\%), while the valid set comprises 11,414 documents (24\%). At the sentence level, the train set of SciDMT contains 738,857 positive sentences (70\%) that contain mentions of DMT entities, while the valid set consists of 314,689 positive sentences (30\%).

\begin{table*}[]
\fontsize{10}{12}\selectfont
\begin{adjustbox}{width=1\linewidth}
\centering
\begin{tabular}{rl}
% i = 173
\hline
GT:& One of the most prominent models of this sort is the \textbf{\color[HTML]{3779D5}{Feature Pyramid Network}} (\textbf{\color[HTML]{3779D5}{FPN}}) proposed by Lin et al. \\
SciBERT:& One of the most prominent models of this sort is the \textbf{\color[HTML]{3779D5}{Feature Pyramid Network}} (\textbf{\color[HTML]{3779D5}{FPN}})  proposed by Lin et al. \\
GPT-3.5:& One of the most prominent models of this sort is the \textbf{\color[HTML]{3779D5}{Feature Pyramid Network (FPN)}} proposed by \textbf{\color[HTML]{3779D5}{Lin et al.}} \\
\hline

% i = 2638
GT:& Note, that the concatenation mode is only relevant for training the \textbf{{\color[HTML]{3779D5}softmax}} classifier.\\
SciBERT: &Note, that the concatenation mode is only relevant for training the \textbf{{\color[HTML]{3779D5}softmax}} classifier.\\
GPT-3.5: &Note, that the concatenation mode is only relevant for \textbf{{\color[HTML]{9E0709}training}} the \textbf{{\color[HTML]{3779D5}softmax classifier}}.\\
\hline

% i=3133
GT:&To achieve such progress, we consider that \textbf{{\color[HTML]{9700FA}Kinetics}} for \textbf{\color[HTML]{3779D5}{3D CNNs}} should be as large-scale as \textbf{{\color[HTML]{9700FA}Ima-geNet}} for \\ &\textbf{\color[HTML]{3779D5}{2D CNNs}}, though no previous work has examined enough about the scale of \textbf{{\color[HTML]{9700FA}Kinetics}}. \\
SciBERT:&To achieve such progress, we consider that Kinetics for 3D CNNs should be as large-scale as Ima-geNet for \\&2D \textbf{\color[HTML]{3779D5}{CNNs}}, though no previous work has examined enough about the scale of Kinetics. \\
GPT-3.5:&To achieve such progress, we consider that \textbf{{\color[HTML]{9700FA}Kinetics}} for \textbf{\color[HTML]{3779D5}{3D CNNs}} should be as large-scale as Ima-geNet for \\ &\textbf{\color[HTML]{3779D5}{2D CNNs}}, though \textbf{{\color[HTML]{9E0709}no previous work}} has examined enough about the scale of \textbf{{\color[HTML]{9700FA}Kinetics}}. \\
\hline

%i=31702 (zero-shot)
GT:& Using \textbf{{\color[HTML]{3779D5}biLMs}} for supervised NLP tasks Given a pre-trained \textbf{{\color[HTML]{3779D5}biLM}} and a supervised architecture for a target \\ &NLP task, it is a simple process to use the \textbf{{\color[HTML]{3779D5}biLM}} to improve the task model.\\
SciBERT:& Using biLMs for supervised NLP tasks Given a pre-trained biLM and a supervised architecture for a target \\ &NLP task, it is a simple process to use the \textbf{{\color[HTML]{3779D5}biLM}} to improve the task model.\\
GPT-3.5:& Using biLMs for \textbf{{\color[HTML]{9E0709}supervised}} NLP tasks Given a pre-trained biLM and a \textbf{{\color[HTML]{9E0709}supervised}} architecture for a \textbf{{\color[HTML]{9E0709}target}} \\ &\textbf{{\color[HTML]{9E0709}NLP task}}, it is a simple process to use the biLM to improve the task model.\\
\hline
 
\end{tabular}
\end{adjustbox}
\caption{Prediction examples for SciBERT and GPT-3.5 on evaluation samples from SciDMT-E. Where the predicted mention tokens are highlighted for 
\textbf{{\color[HTML]{9700FA}dataset (D)}}, \textbf{{\color[HTML]{3779D5}method (M)}} and \textbf{{\color[HTML]{9E0709}task (T)}}.}
\label{tab:pred_samples} % UPDATED  06/14
\end{table*}
\section{Experimental Results}
In this section, we present the experimental specifications and report the results of our experiments. The evaluation is conducted on the manually annotated evaluation sets: SciDMT-E and SciREX-E. The average and standard deviation of performance scores, including F\textsubscript{1}, precision (P), and recall (R), are calculated based on the exact scores.

\subsection{Baselines Evaluation}\label{sec:SOTA-results}
The NER models discussed in Section~\ref{sec:ner_models}, undergo 3 rounds of training using randomly shuffled training sets from SciDMT, except GPT-3.5, which is evaluated for 1 round.  Performance scores are computed on SciDMT-E, SciREX-E, and various subsets of SciDMT-E. The results are summarized in Table~\ref{tab:sotaner} and prediction samples are shown in Table~\ref{tab:pred_samples}.

The performance on the two evaluation sets, SciREX-E and SciDMT-E, is similar, indicating comparable dataset difficulty. Surprisingly, for BERT and SciBERT, the performance of the unseen subset in SciDMT-E is higher than the overall average performance, contrary to our expectations. This may be due to the limited sample size, which might have excluded more challenging cases. Additionally, we observe that dataset mentions are generally easier to detect compared to method and task mentions, possibly because dataset names are more standardized and have fewer naming variations.

SciBERT and BERT exhibit similar performances among the different models, achieving the highest overall performance. The variations of BiLSTM using Word2Vec embedding (BiLSTM-W) and GloVe embedding (BiLSTM-G) perform similarly to the original BiLSTM, without notable improvements. This aligns with previous research that has shown Word2Vec and GloVe to be equivalent in terms of module hyperparameter tuning \cite{levy_goldberg_dagan_2015, arora_li_liang_ma_risteski_2016}. Without sophisticated feature learning, CRF does not perform as competitively as the other models.

GPT-3.5 without fine-tuning achieves slightly lower scores compared to the trained models, but still demonstrates knowledge about scientific entities without explicit learning. It predicts some of the general concept words (e.g.: `training', `inference', and `modification') and citations (e.g.: `Fortunato et al. 2017') as scientific entities, which are not included in our manual evaluation sets. We hypothesize that as GPT-3.5 is trained with Common Crawl and scientific papers often have web presences, GPT-3.5 may have read many scientific papers during training and accumulated knowledge about understanding and identifying scientific entities. By including SciDMT in its training or employing few-shot learning, the performance of GPT-3.5 can potentially be further improved. However, GPT-3.5 struggles with isolating the correct entity from descriptions or strings with multiple mentions, as shown in Table~\ref{tab:pred_samples}. Additionally, like other trained models, GPT-3.5 encounters difficulties in recognizing mentions with uncommon dash patterns, such as `Ima-geNet'.

\subsection{Error Analysis} \label{sec:error_analysis}
Based on the performance of the best model, SciBERT, we conduct an error analysis to identify common patterns among erroneous instances. These patterns include long sequences with more than 200 characters, sequences with multiple mentions, and sequences with unseen mentions. 

Here, 'unseen mentions' are twofold: firstly, they include the annotated unseen entity mentions previously discussed. Secondly, they encompass mentions identified by human annotators that could not be linked to the SciDMT dictionary. Like the annotated unseen entities, these unseen mentions lack any representation in the training dataset. 

We compute the number of evaluated sentences exhibiting each pattern and their corresponding performance scores in Table~\ref{tab:error}.  SciBERT demonstrates below-average performance across all common patterns, with the lowest performance observed in the unseen category. 

In Table~\ref{tab:pred_samples}, the last two examples are samples demonstrate cases of long sequences and multiple mentions, while the last example is an sample for unseen mentions as it is the one containing unseen entity `biLM'.

\begin{table}[]

\centering
\begin{adjustbox}{width=1\linewidth}

\begin{tabular}{r|c|ccc}
\hline
Eval. Group & N & F1 & P & R \\ \hline
All & 17,053 & .649 & .531 & .833 \\
Long Sequences & 1,670 & .547 & .420 & .786 \\
Multiple Mentions & 8,312 & .577 & .440 & .836 \\
Unseen Mentions & 4,212 & .529 & .423 & .708 \\ \hline
\end{tabular}
\end{adjustbox}
\caption{Error analysis for SciBERT. N represents the number of evaluated sequences with different features.}
\label{tab:error} % UPDATED  06/14
% \vspace{-10pt}
\end{table}

\begin{figure*}[]
\includegraphics[width=1\linewidth]{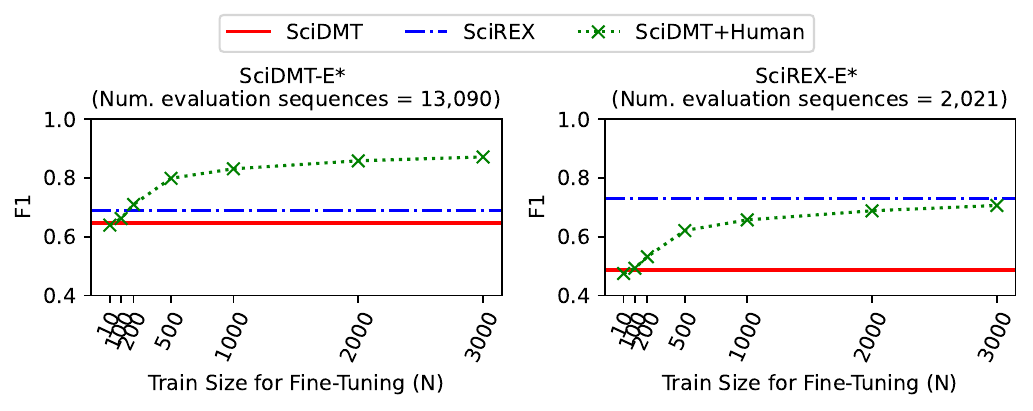}
\caption{Trend of F1 when varying the number (N) of human-annotated samples used for fine-tuning. Each line in the graph, represented in the legend, corresponds to a model being trained with a distinct dataset.}
\label{fig:finetune} %UPDATED 06/15

\end{figure*}

\subsection{Fine-Tuning with Human Labels}

To assess the effectiveness of SciDMT as a training resource, we compare SciBERT models trained solely with weak labels from SciDMT to those trained solely with human-annotated labels (human labels) from SciREX. We also investigate the minimum number of human labels needed to achieve a comparable level of performance. For the fine-tuning training set, we randomly sample 1500 positive sequences from each of our human-annotated evaluation sets: SciDMT-E and SciREX-E, while retaining the remaining sequences as the fine-tune evaluation sets (SciDMT-E* and SciREX-E*).

We develop three types of SciBERT models:
\begin{itemize}
    \item \(M_{SciDMT}\), which is trained using the weak labels from SciDMT.
    \item \(M_{SciREX}\), which is trained using the human labels from SciREX, comprising only human-annotated samples. The training set of SciREX consists of 60,021 positive sequences, excluding those that overlap with SciDMT's valid set and our human-annotated evaluation sets.
    \item \(M_{SciDMT+N}\), which is fine-tuned on top of \(M_{SciDMT}\) with N randomly sampled human labels from the fine-tuning training set. We experiment with different values of N, including 10, 100, 200, 500, 1000, 2000, and 3000.
\end{itemize}

All models are evaluated separately on SciDMT-E* and SciREX-E*. The models' performance and the trend in F1 scores as the number of human annotations varies are shown in Figure~\ref{fig:finetune}. 

As anticipated, \(M_{SciDMT}\) trained solely with weak labels performs lower \(M_{SciREX}\) trained with human labels. In terms of fine-tuning, \(M_{SciDMT+N}\) achieves better performance than \(M_{SciDMT}\) with 100 human labels. 

On SciDMT-E*, \(M_{SciDMT+N}\) surpasses the performance of \(M_{SciREX}\) with only 200 human labels, outperforming the model trained with 60K human labels. Moreover, fine-tuning with 3,000 human-annotated sequences further improves the performance, achieving 0.88 F1 scores on SciDMT-E*.

On SciREX-E*, where \(M_{SciREX}\) has the advantage of being trained in the same domain, \(M_{SciDMT+N}\) needs 3000 human labels to achieve similar performance to \(M_{SciREX}\). In other words, fine-tuning the pre-trained model from SciDMT with approximately 10 human-annotated documents yields comparable performance to the model trained with around 245 documents.

\begin{figure}[]
\centering
\includegraphics[width=1\linewidth]{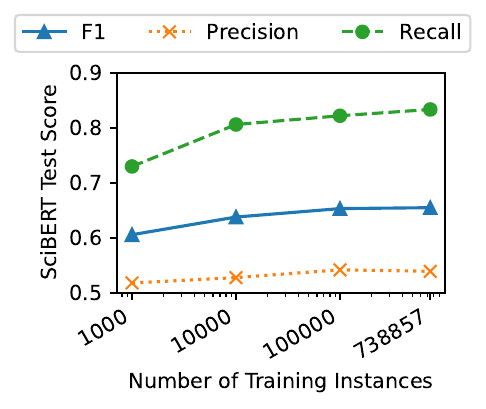}
\caption{Validation performance of SciBERT when training on SciDMT as the train size increases.}
\label{fig:trainsize} %UPDATED 06/14
\end{figure}

\subsection{Impact of Training Scale on Performance} \label{sec:exp-results-train-size} 
As part of our ablation study, we investigate the training benefits derived from the large size of SciDMT. We train SciBERT using different training set sizes, randomly sampled from the entire training data: 10\textsuperscript{3}, 10\textsuperscript{4}, 10\textsuperscript{5}, and the complete training data consisting of 739K samples. The performance scores on SciDMT-E are plotted using a logarithmic scale in Figure~\ref{fig:trainsize}.

Our analysis reveals that the most significant improvement in model performance occurs when increasing the training size from 1,000 to 10,000 sequences. The recall score continues to improve as the training size increases, while the F1 score and precision remain relatively stable beyond a training size of 100,000. This suggests that the model tends to predict more false positives with larger training sizes. 

To better leverage the large size and diversity of mentions in SciDMT and further enhance the model's performance, it can be beneficial to balance the training datasets by sampling biased toward challenging cases. This strategy can focus on samples with common features observed in the error analysis (Section~\ref{sec:error_analysis}).

\section{Limitations}
SciDMT is a large-scale corpus annotated through distant supervision. This approach sacrifices accuracy for scale. The current scope of SciDMT is limited to scientific mentions that can be linked to the SciDMT dictionary, resulting in missing labels for scientific mentions that are not listed on PwC websites or that have variations not included in the regular expression. This limitation may introduce annotation noise, especially when dealing with subversions that are not explicitly listed in PwC. In addition, SciDMT may inadvertently inherit biases from its primary source, PwC's data. This reliance could lead to disproportionate emphasis or neglect of certain topics within the corpus.

Furthermore, SciDMT does not include annotations for ambiguous cases, where distinct entities have the same name or share acronyms, nor does it consider changes in naming conventions over time. Similar limitations apply to other corpora created using distant supervision, as annotation accuracy heavily relies on manual correction. 

Additionally, SciDMT does not annotate pronominal references to entities, resulting in incomplete coreference information compared to corpora like SciERC. Despite these limitations, the large-scale data obtained through distant supervision proves valuable for training deep learning models, a sentiment echoed in previous studies \citep{emonet,SchneiderHCMD22,su2019} and our experimental findings in Section~\ref{sec:exp-results-train-size}.

\section{Conclusion \& Future Directions}
We presented SciDMT, the largest corpus specifically created for the study of scientific entity mention detection (SEMD). SciDMT offers a substantial size, diverse entity mentions, and comprehensive entity-linking information, making it a valuable resource and a benchmark for the development and evaluation of advanced scientific information extraction models. 

The experiments conducted using various NER models on SciDMT provide valuable insights and performance baselines for SEMD. The error analysis conducted sheds light on the existing challenges and unveils opportunities for innovation in SEMD.

Moving forward, our focus is on the iterative enhancement of SciDMT. We aim to augment the corpus by broadening the spectrum of annotated entities, refining weak labels, and increasing the corpus size. The incorporation of sophisticated post-processing techniques~\cite{DragutWYSM12,DragutWSYM15,SchneiderD15,RE_distant} to cleanse distant supervision labels is also on our agenda. Additionally, future work can focus on addressing more challenging instances, such as unseen and ambiguous mentions, to further enhance the performance of scientific mention detection models.

In conclusion, SciDMT presents a significant contribution to the field of SEMD by providing a large-scale corpus and performance baselines for SEMD models. We hope SciDMT will inspire and drive future research in scientific information extraction.

\section{Acknowledgement}
This work was supported by the National Science Foundation awards III-2107213, III-2107518, and ITE-2333789. We also thank undergraduate students Eric Reizas and Elle Nguyen at Temple University for their valuable contributions to our project.

\section{Ethics Statement}
We use public data from scientific documents for our experiments.

\nocite{*}
\section{Bibliographical References}\label{sec:reference}

\bibliographystyle{lrec-coling2024-natbib}
\bibliography{main_references}

\begin{thebibliography}{15}
\expandafter\ifx\csname natexlab\endcsname\relax\def\natexlab#1{#1}\fi

\bibitem[{Augenstein et~al.(2017)Augenstein, Das, Riedel, Vikraman, and
  McCallum}]{semeval2017}
Augenstein, Isabelle and Das, Mrinal and Riedel, Sebastian and Vikraman,
  Lakshmi and McCallum, Andrew. 2017.
\newblock \emph{{S}em{E}val 2017 Task 10: {S}cience{IE} - Extracting Keyphrases
  and Relations from Scientific Publications}.
\newblock PID
  \href{https://github.com/UKPLab/semeval2017-scienceie}{https://github.com/UKPLab/semeval2017-scienceie}.

\bibitem[{Blagec et~al.(2021)Blagec, Barbosa-Silva, Ott, and Samwald}]{ITO}
Kathrin Blagec and Adriano Barbosa-Silva and Simon Ott and Matthias Samwald.
  2021.
\newblock \emph{A curated, ontology-based, large-scale knowledge graph of
  artificial intelligence tasks and benchmarks}.
\newblock PID
  \href{https://github.com/OpenBioLink/ITO}{https://github.com/OpenBioLink/ITO}.

\bibitem[{Duck et~al.(2013)Duck, Nenadic, Brass, Robertson, and
  Stevens}]{bioNerDS}
Duck, Geraint and Nenadic, Goran and Brass, Andy and Robertson, David L and
  Stevens, Robert. 2013.
\newblock \href {https://doi.org/10.1186/1471-2105-14-194} {\emph{bioNerDS:
  exploring bioinformatics’ database and software use through literature
  mining}}.
\newblock PID
  \href{http://bionerds.sourceforge.net/}{http://bionerds.sourceforge.net/}.

\bibitem[{F{\"a}rber et~al.(2021)F{\"a}rber, Albers, and
  Sch{\"u}ber}]{farber2021identifying}
F{\"a}rber, Michael and Albers, Alexander and Sch{\"u}ber, Felix. 2021.
\newblock \emph{Identifying Used Methods and Datasets in Scientific
  Publications}.
\newblock PID \href{https://github.
  com/michaelfaerber/scholarly-entity-usage-detection}{https://github.
  com/michaelfaerber/scholarly-entity-usage-detection}.

\bibitem[{G{\'a}bor et~al.(2018)G{\'a}bor, Buscaldi, Schumann, QasemiZadeh,
  Zargayouna, and Charnois}]{semeval2018}
G{\'a}bor, Kata and Buscaldi, Davide and Schumann, Anne-Kathrin and
  QasemiZadeh, Behrang and Zargayouna, Ha{\"\i}fa and Charnois, Thierry. 2018.
\newblock \href {https://doi.org/10.18653/v1/S18-1111} {\emph{{S}em{E}val-2018
  Task 7: Semantic Relation Extraction and Classification in Scientific
  Papers}}.
\newblock Association for Computational Linguistics.
\newblock PID \href{https://competitions.codalab.org/competitions/
  17422}{https://competitions.codalab.org/competitions/ 17422}.

\bibitem[{Heddes et~al.(2021)Heddes, Meerdink, Pieters, and Marx}]{heddes2021}
Heddes, Jenny and Meerdink, Pim and Pieters, Miguel and Marx, Maarten. 2021.
\newblock \href {https://doi.org/10.3390/data6080084} {\emph{The Automatic
  Detection of Dataset Names in Scientific Articles}}.
\newblock PID
  \href{https://github.com/xjaeh/ner\_dataset\_recognition}{https://github.com/xjaeh/ner\_dataset\_recognition}.

\bibitem[{Hou et~al.(2019)Hou, Jochim, Gleize, Bonin, and Ganguly}]{NLP-TDMS}
Hou, Yufang and Jochim, Charles and Gleize, Martin and Bonin, Francesca and
  Ganguly, Debasis. 2019.
\newblock \href {https://doi.org/10.18653/v1/P19-1513} {\emph{Identification of
  Tasks, Datasets, Evaluation Metrics, and Numeric Scores for Scientific
  Leaderboards Construction}}.
\newblock Association for Computational Linguistics.
\newblock PID
  \href{https://github.com/IBM/science-result-extractor}{https://github.com/IBM/science-result-extractor}.

\bibitem[{Hou et~al.(2021)Hou, Jochim, Gleize, Bonin, and Ganguly}]{tdmsci}
Hou, Yufang and Jochim, Charles and Gleize, Martin and Bonin, Francesca and
  Ganguly, Debasis. 2021.
\newblock \href {https://doi.org/10.18653/v1/2021.eacl-main.59}
  {\emph{{TDMS}ci: A Specialized Corpus for Scientific Literature Entity
  Tagging of Tasks Datasets and Metrics}}.
\newblock Association for Computational Linguistics.
\newblock PID
  \href{https://github.com/IBM/science-result-extractor}{https://github.com/IBM/science-result-extractor}.

\bibitem[{Jain et~al.(2020)Jain, van Zuylen, Hajishirzi, and Beltagy}]{scirex}
Jain, Sarthak and van Zuylen, Madeleine and Hajishirzi, Hannaneh and Beltagy,
  Iz. 2020.
\newblock \href {https://doi.org/10.18653/v1/2020.acl-main.670}
  {\emph{{S}ci{REX}: {A} Challenge Dataset for Document-Level Information
  Extraction}}.
\newblock Association for Computational Linguistics.
\newblock PID
  \href{https://github.com/allenai/SciREX}{https://github.com/allenai/SciREX}.

\bibitem[{Lo et~al.(2020)Lo, Wang, Neumann, Kinney, and Weld}]{S2ORC}
Lo, Kyle and Wang, Lucy Lu and Neumann, Mark and Kinney, Rodney and Weld,
  Daniel. 2020.
\newblock \href {https://doi.org/10.18653/v1/2020.acl-main.447}
  {\emph{{S}2{ORC}: The Semantic Scholar Open Research Corpus}}.
\newblock Association for Computational Linguistics.
\newblock PID
  \href{https://github.com/allenai/s2orc}{https://github.com/allenai/s2orc}.

\bibitem[{Luan et~al.(2018)Luan, He, Ostendorf, and Hajishirzi}]{SCIERC}
Luan, Yi and He, Luheng and Ostendorf, Mari and Hajishirzi, Hannaneh. 2018.
\newblock \emph{Multi-Task Identification of Entities, Relations, and
  Coreference for Scientific Knowledge Graph Construction}.

\bibitem[{Pan et~al.(2023)Pan, Zhang, Dragut, Caragea, and Latecki}]{DMDD}
Pan, Huitong and Zhang, Qi and Dragut, Eduard and Caragea, Cornelia and
  Latecki, Longin Jan. 2023.
\newblock \href {https://doi.org/10.1162/tacl_a_00592} {\emph{{DMDD: A
  Large-Scale Dataset for Dataset Mentions Detection}}}.
\newblock Transactions of the Association for Computational Linguistics.

\bibitem[{Tabassum et~al.(2020)Tabassum, Xu, and
  Ritter}]{tabassum-etal-2020-wnut}
Tabassum, Jeniya and Xu, Wei and Ritter, Alan. 2020.
\newblock \href {https://doi.org/10.18653/v1/2020.wnut-1.33} {\emph{{WNUT}-2020
  Task 1 Overview: Extracting Entities and Relations from Wet Lab Protocols}}.
\newblock Association for Computational Linguistics.
\newblock PID
  \href{https://github.com/jeniyat/WNUT\_2020\_NER}{https://github.com/jeniyat/WNUT\_2020\_NER}.

\bibitem[{{Yao} et~al.(2019){Yao}, {Hou}, {Ye}, {Wu}, {Zhang}, and {Wu}}]{MDER}
{Yao}, Rujing and {Hou}, Linlin and {Ye}, Yingchun and {Wu}, Ou and {Zhang}, Ji
  and {Wu}, Jian. 2019.
\newblock \href {http://arxiv.org/abs/1911.13096} {\emph{{Method and Dataset
  Mining in Scientific Papers}}}.
\newblock PID
  \href{https://arxiv.org/pdf/1911.13096.pdf}{https://arxiv.org/pdf/1911.13096.pdf}.

\bibitem[{Zhao et~al.(2019)Zhao, Luo, Feng, Zheng, and Liu}]{SciRes}
Zhao, He and Luo, Zhunchen and Feng, Chong and Zheng, Anqing and Liu, Xiaopeng.
  2019.
\newblock \href {https://doi.org/10.18653/v1/D19-1524} {\emph{A Context-based
  Framework for Modeling the Role and Function of On-line Resource Citations in
  Scientific Literature}}.
\newblock Association for Computational Linguistics.
\newblock PID
  \href{https://github.com/zhaohe1995/SciRes}{https://github.com/zhaohe1995/SciRes}.

\end{thebibliography}


\begin{thebibliography}{26}
\expandafter\ifx\csname natexlab\endcsname\relax\def\natexlab#1{#1}\fi

\bibitem[{Abdul-Mageed and Ungar(2017)}]{emonet}
Muhammad Abdul-Mageed and Lyle Ungar. 2017.
\newblock {E}mo{N}et: Fine-grained emotion detection with gated recurrent
  neural networks.
\newblock In \emph{Association for Computational Linguistics (ACL)}, pages
  718--728, Vancouver, Canada.

\bibitem[{Alshehri et~al.(2022)Alshehri, Stanojevic, Dragut, and
  Obradovic}]{AlshehriSDO22}
Jumanah Alshehri, Marija Stanojevic, Eduard Dragut, and Zoran Obradovic. 2022.
\newblock On label quality in class imbalance setting -a case study.
\newblock In \emph{ICMLA}, pages 1666--1671.

\bibitem[{Arora et~al.(2016)Arora, Li, Liang, Ma, and
  Risteski}]{arora_li_liang_ma_risteski_2016}
Sanjeev Arora, Yuanzhi Li, Yingyu Liang, Tengyu Ma, and Andrej Risteski. 2016.
\newblock \href {https://doi.org/10.1162/tacl_a_00106} {A latent variable model
  approach to pmi-based word embeddings}.
\newblock \emph{Transactions of the Association for Computational Linguistics},
  4:385–399.

\bibitem[{Beltagy et~al.(2019)Beltagy, Lo, and
  Cohan}]{beltagy-etal-2019-scibert}
Iz~Beltagy, Kyle Lo, and Arman Cohan. 2019.
\newblock \href {https://www.aclweb.org/anthology/D19-1371} {Scibert: A
  pretrained language model for scientific text}.
\newblock In \emph{Empirical Methods in Natural Language Processing (EMNLP)}.
  Association for Computational Linguistics.

\bibitem[{Bird et~al.(2009)Bird, Klein, and Loper}]{10.5555/1717171}
Steven Bird, Ewan Klein, and Edward Loper. 2009.
\newblock \emph{Natural Language Processing with Python}, 1st edition.
\newblock O'Reilly Media, Inc.

\bibitem[{Bose et~al.(2021)Bose, Srinivasan, Sleeman, Palta, Kapoor, and
  Ghosh}]{bose_2021}
Priyankar Bose, Sriram Srinivasan, William~C. Sleeman, Jatinder Palta, Rishabh
  Kapoor, and Preetam Ghosh. 2021.
\newblock \href {https://doi.org/10.3390/app11188319} {A survey on recent named
  entity recognition and relationship extraction techniques on clinical texts}.
\newblock \emph{Applied Sciences}, 11(18):8319.

\bibitem[{Davies and Fleiss(1982)}]{measure_agreement}
Mark Davies and Joseph~L. Fleiss. 1982.
\newblock \href {http://www.jstor.org/stable/2529886} {Measuring agreement for
  multinomial data}.
\newblock \emph{Biometrics}, 38(4):1047--1051.

\bibitem[{Devlin et~al.(2018)Devlin, Chang, Lee, and
  Toutanova}]{bert_base_cased}
Jacob Devlin, Ming{-}Wei Chang, Kenton Lee, and Kristina Toutanova. 2018.
\newblock \href {http://arxiv.org/abs/1810.04805} {{BERT:} pre-training of deep
  bidirectional transformers for language understanding}.
\newblock \emph{CoRR}, abs/1810.04805.

\bibitem[{Dragut et~al.(2015)Dragut, Wang, Sistla, Yu, and Meng}]{DragutWSYM15}
Eduard~C. Dragut, Hong Wang, A.~Prasad Sistla, Clement~T. Yu, and Weiyi Meng.
  2015.
\newblock Polarity consistency checking for domain independent sentiment
  dictionaries.
\newblock \emph{TKDE}, 27(3):838--851.

\bibitem[{Dragut et~al.(2012)Dragut, Wang, Yu, Sistla, and Meng}]{DragutWYSM12}
Eduard~C. Dragut, Hong Wang, Clement~T. Yu, A.~Prasad Sistla, and Weiyi Meng.
  2012.
\newblock Polarity consistency checking for sentiment dictionaries.
\newblock In \emph{ACL}, pages 997--1005.

\bibitem[{Gutierrez et~al.(2022)Gutierrez, McNeal, Washington, Chen, Li, Sun,
  and Su}]{Gutierrez2022ThinkingAG}
Bernal~Jimenez Gutierrez, Nikolas McNeal, Clay Washington, You Chen, Lang Li,
  Huan Sun, and Yu~Su. 2022.
\newblock Thinking about gpt-3 in-context learning for biomedical ie? think
  again.
\newblock \emph{ArXiv}, abs/2203.08410.

\bibitem[{Levy et~al.(2015)Levy, Goldberg, and
  Dagan}]{levy_goldberg_dagan_2015}
Omer Levy, Yoav Goldberg, and Ido Dagan. 2015.
\newblock \href {https://doi.org/10.1162/tacl_a_00134} {Improving
  distributional similarity with lessons learned from word embeddings}.
\newblock \emph{Transactions of the Association for Computational Linguistics},
  3:211–225.

\bibitem[{Mikolov et~al.(2013)Mikolov, Chen, Corrado, and Dean}]{word2vec}
Tomas Mikolov, Kai Chen, Greg~S. Corrado, and Jeffrey Dean. 2013.
\newblock \href {http://arxiv.org/abs/1301.3781} {Efficient estimation of word
  representations in vector space}.

\bibitem[{Mintz et~al.(2009)Mintz, Bills, Snow, and Jurafsky}]{mintz}
Mike Mintz, Steven Bills, Rion Snow, and Daniel Jurafsky. 2009.
\newblock \href {https://aclanthology.org/P09-1113} {Distant supervision for
  relation extraction without labeled data}.
\newblock In \emph{Proceedings of the Joint Conference of the 47th Annual
  Meeting of the {ACL} and the 4th International Joint Conference on Natural
  Language Processing of the {AFNLP}}, pages 1003--1011, Suntec, Singapore.
  Association for Computational Linguistics.

\bibitem[{Neumann et~al.(2019)Neumann, King, Beltagy, and
  Ammar}]{neumann-etal-2019-scispacy}
Mark Neumann, Daniel King, Iz~Beltagy, and Waleed Ammar. 2019.
\newblock \href {https://doi.org/10.18653/v1/W19-5034} {{S}cispa{C}y: {F}ast
  and {R}obust {M}odels for {B}iomedical {N}atural {L}anguage {P}rocessing}.
\newblock In \emph{Proceedings of the 18th BioNLP Workshop and Shared Task},
  pages 319--327, Florence, Italy. Association for Computational Linguistics.

\bibitem[{Ntroduction(2005)}]{ace2005}
Ii.~I Ntroduction. 2005.
\newblock The ace 2005 (ace 05) evaluation plan evaluation of the detection and
  recognition of ace entities, values, temporal expressions, relations, and
  events.

\bibitem[{Ouyang et~al.(2022)Ouyang, Wu, Jiang, Almeida, Wainwright, Mishkin,
  Zhang, Agarwal, Slama, Ray, Schulman, Hilton, Kelton, Miller, Simens, Askell,
  Welinder, Christiano, Leike, and Lowe}]{instruct_gpt}
Long Ouyang, Jeff Wu, Xu~Jiang, Diogo Almeida, Carroll~L Wainwright, Pamela
  Mishkin, Chong Zhang, Sandhini Agarwal, Katarina Slama, Alex Ray, John
  Schulman, Jacob Hilton, Fraser Kelton, Luke Miller, Maddie Simens, Amanda
  Askell, Peter Welinder, Paul Christiano, Jan Leike, and Ryan Lowe. 2022.
\newblock \href {https://arxiv.org/abs/2203.02155} {Training language models to
  follow instructions with human feedback}.

\bibitem[{Pennington et~al.(2014)Pennington, Socher, and Manning}]{glove}
Jeffrey Pennington, Richard Socher, and Christopher Manning. 2014.
\newblock {G}lo{V}e: Global vectors for word representation.
\newblock In \emph{Empirical Methods in Natural Language Processing (EMNLP)}.

\bibitem[{Schneider et~al.(2022)Schneider, He, Chen, Mukherjee, and
  Dragut}]{SchneiderHCMD22}
Andrew Schneider, Lihong He, Zhijia Chen, Arjun Mukherjee, and Eduard Dragut.
  2022.
\newblock {COIN} {--} an inexpensive and strong baseline for predicting out of
  vocabulary word embeddings.
\newblock In \emph{COLING}, pages 3984--3993.

\bibitem[{Schneider and Dragut(2015)}]{SchneiderD15}
Andrew~T. Schneider and Eduard~C. Dragut. 2015.
\newblock Towards debugging sentiment lexicons.
\newblock In \emph{ACL}, pages 1024--1034.

\bibitem[{Smirnova and Cudré-Mauroux(2018)}]{RE_distant}
Alisa Smirnova and Philippe Cudré-Mauroux. 2018.
\newblock \href {https://doi.org/https://doi.org/10.1145/3241741} {Relation
  extraction using distant supervision}.
\newblock \emph{ACM Computing Surveys}, 51(5):1–35.

\bibitem[{Stenetorp et~al.(2012)Stenetorp, Pyysalo, Topi\'{c}, Ohta, Ananiadou,
  and Tsujii}]{brat}
Pontus Stenetorp, Sampo Pyysalo, Goran Topi\'{c}, Tomoko Ohta, Sophia
  Ananiadou, and Jun'ichi Tsujii. 2012.
\newblock {brat}: a web-based tool for {NLP}-assisted text annotation.
\newblock In \emph{Proceedings of the Demonstrations Session at {EACL} 2012},
  Avignon, France. Association for Computational Linguistics.

\bibitem[{Su et~al.(2019)Su, Li, Wu, and Vijay-Shanker}]{su2019}
Peng Su, Gang Li, Cathy Wu, and K.~Vijay-Shanker. 2019.
\newblock \href {https://doi.org/10.1371/journal.pone.0216913} {Using distant
  supervision to augment manually annotated data for relation extraction}.
\newblock \emph{PLOS ONE}, 14(7):e0216913.

\bibitem[{Tang et~al.(2008)Tang, Zhang, Yao, Li, Zhang, and
  Su}]{10.1145/1401890.1402008}
Jie Tang, Jing Zhang, Limin Yao, Juanzi Li, Li~Zhang, and Zhong Su. 2008.
\newblock \href {https://doi.org/10.1145/1401890.1402008} {Arnetminer:
  Extraction and mining of academic social networks}.
\newblock In \emph{Proceedings of the 14th ACM SIGKDD International Conference
  on Knowledge Discovery and Data Mining}, KDD '08, page 990–998, New York,
  NY, USA. Association for Computing Machinery.

\bibitem[{Zhang et~al.(2019)Zhang, He, Dragut, and Vucetic}]{zhang2019invest}
Shanshan Zhang, Lihong He, Eduard Dragut, and Slobodan Vucetic. 2019.
\newblock How to invest my time: Lessons from human-in-the-loop entity
  extraction.
\newblock In \emph{Proceedings of the 25th ACM SIGKDD International Conference
  on Knowledge Discovery \& Data Mining}, pages 2305--2313.

\bibitem[{Zhang et~al.(2018)Zhang, He, Vucetic, and Dragut}]{ZhangHVD18}
Shanshan Zhang, Lihong He, Slobodan Vucetic, and Eduard Dragut. 2018.
\newblock Regular expression guided entity mention mining from noisy web data.
\newblock In \emph{EMNLP}, pages 1991--2000.

\end{thebibliography}

\section{Language Resource References}
\label{lr:ref}
\bibliographystylelanguageresource{lrec-coling2024-natbib}
\bibliographylanguageresource{main_language_resource}

\end{document}